\documentclass[letterpaper]{article} 
\usepackage{aaai2026}  
\usepackage{times}  
\usepackage{helvet}  
\usepackage{courier}  
\usepackage[hyphens]{url}  
\usepackage{graphicx} 
\usepackage{natbib}  
\usepackage{caption} 
\usepackage{algorithm}
\usepackage{algorithmicx}
\usepackage{algpseudocode}
\usepackage{amsmath,amsfonts}
\usepackage{newfloat}
\usepackage{listings}
\DeclareCaptionStyle{ruled}{labelfont=normalfont,labelsep=colon,strut=off} 
\floatstyle{ruled}
\newfloat{listing}{tb}{lst}{}
\floatname{listing}{Listing}
\title{Instance Generation for Meta-Black-Box Optimization \\ through Latent Space Reverse Engineering}
\author{
    Chen Wang\textsuperscript{\rm 1},
    Yue-Jiao Gong\textsuperscript{\rm 1},
    Zhiguang Cao\textsuperscript{\rm 2},
    Zeyuan Ma\textsuperscript{\rm 1}\thanks{Corresponding author~(scut.crazynicolas@gmail.com)}
}
\affiliations{
    \textsuperscript{\rm 1}South China University of Technology\\
    \textsuperscript{\rm 2}Singapore Management University

}

\begin{document}

\maketitle

\begin{abstract}
To relieve intensive human-expertise required to design optimization algorithms, recent Meta-Black-Box Optimization (MetaBBO) researches leverage generalization strength of meta-learning to train neural network-based algorithm design policies over a predefined training problem set, which automates the adaptability of the low-level optimizers on unseen problem instances. Currently, a common training problem set choice in existing MetaBBOs is well-known benchmark suites CoCo-BBOB. Although such choice facilitates the MetaBBO's development, problem instances in CoCo-BBOB are more or less limited in diversity, raising the risk of overfitting of MetaBBOs, which might further results in poor generalization. In this paper, we propose an instance generation approach, termed as \textbf{LSRE}, which could generate diverse training problem instances for MetaBBOs to learn more generalizable policies. LSRE first trains an autoencoder which maps high-dimensional problem features into a 2-dimensional latent space. Uniform-grid sampling in this latent space leads to hidden representations of problem instances with sufficient diversity. By leveraging a genetic-programming approach to search function formulas with minimal L2-distance to these hidden representations, LSRE reverse engineers a diversified problem set, termed as \textbf{Diverse-BBO}. We validate the effectiveness of LSRE by training various MetaBBOs on Diverse-BBO and observe their generalization performances on either synthetic or realistic scenarios. Extensive experimental results underscore the superiority of Diverse-BBO to existing training set choices in MetaBBOs. Further ablation studies not only demonstrate the effectiveness of design choices in LSRE, but also reveal interesting insights on instance diversity and MetaBBO's generalization. We provide the code of LSRE and Diverse-BBO at \url{https://github.com/MetaEvo/Diverse-BBO}. 
\end{abstract}


\section{Introduction}
Black-Box Optimization (BBO) problems, with neither the objective function nor corresponding derivative is accessible, are challenging to solve. Although Evolutionary Computation (EC) methods, such as Genetic Algorithm (GA), Particle Swarm Optimization (PSO), Differential Evolution (DE), as well as their modern (self-)adaptive variants are widely recognized as effective solutions, they often require tuning or even re-design for novel scenario adaption.


To automate such manual algorithm design tasks, recent Meta-Black-Box Optimization~(MetaBBO) researches propose learning generalizable design policy through meta-learning~\cite{metabox,ma2025toward}. MetaBBO naturally forms a bi-level algorithmic structure. At the meta level, a learnable algorithm design policy~(e.g., a neural network) outputs flexible design choices~(e.g., parameter control~\cite{GLEET,ConfigX} and algorithm selection~\cite{RLDAS}) for low-level optimizer. At the lower level, a BBO optimizer is deployed to optimize target problem with the given design choice. The meta-objective is to train a design policy that enhances low-level BBO process. Since such paradigm provides a promising way to learn automated optimization algorithm design, MetaBBO has been widely explored and discussed in recent two years~\cite{rl-surrogate,DesignX,kl-lsgo,huang2025evaluation,metade,mo2025autosgnn}.

\begin{figure}[t]
    \centering
    \includegraphics[width=0.80\linewidth]{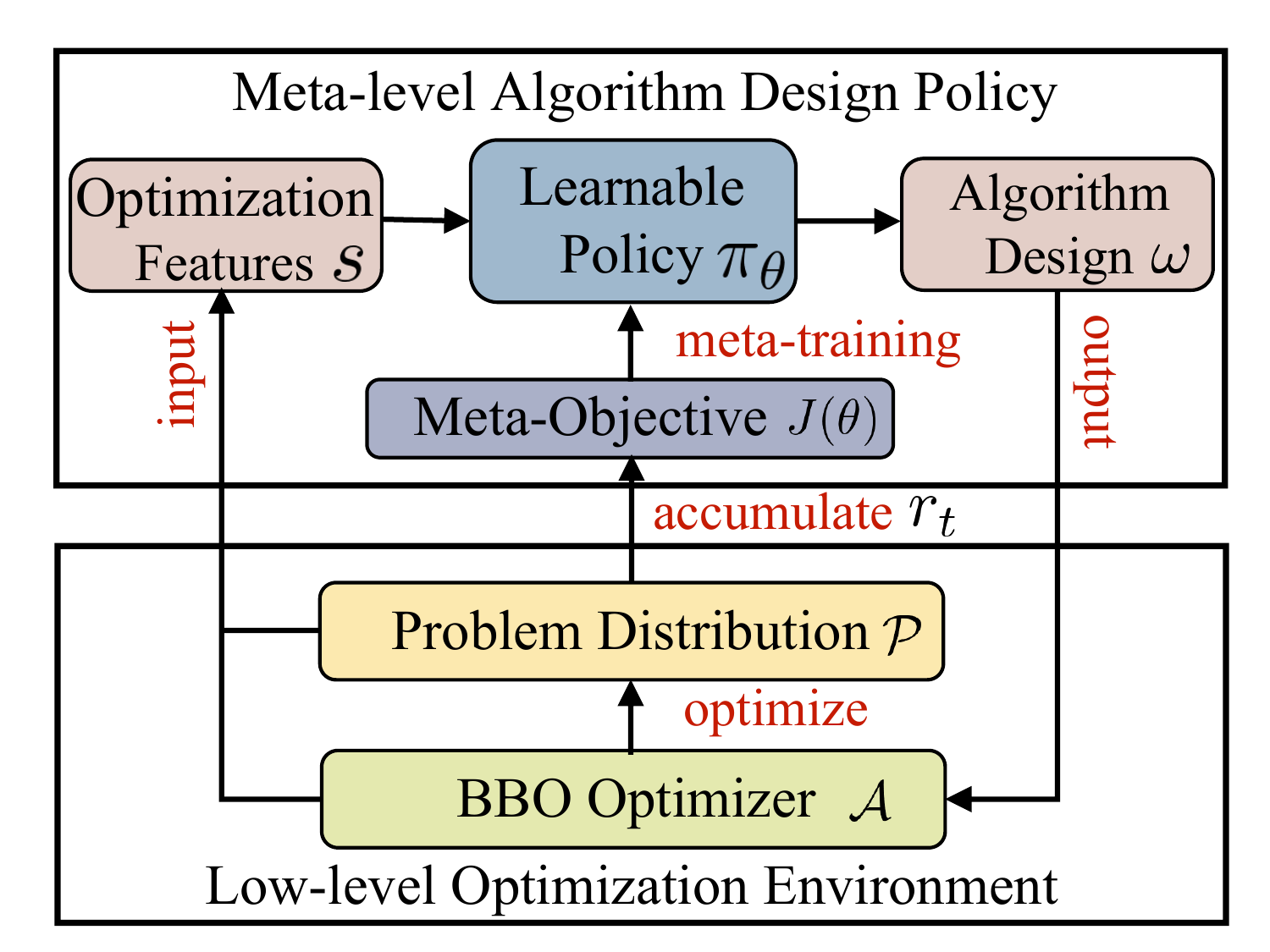}
    \caption{General workflow of MetaBBO approaches.}
    \label{fig:intro}
\end{figure}

Given that MetaBBO essentially follows meta-learning paradigm~\cite{meta-learning-1,meta-learning}, it unavoidably requires domain generalization designs~\cite{meta-dataset-1}, i.e., meta-augmentation of training data~\cite{meta-dataset}, so as to ensure the adaption during the testing time. To be specific, as shown in the yellow box of Fig.~\ref{fig:intro}, existing MetaBBO approaches typically train their policies over a training problem set. The diversity of this set assures that the meta-level policy generalizes across diverse optimization problem characteristics~\cite{neurela}. Common choices of the problem set in existing MetaBBO approaches are synthetic benchmarks such as CoCo-BBOB~\cite{bbob2010}. Although these representative problem instances hold a long-standing history and reputation in the evaluation of BBO methods, they are primarily hand-crafted to differentiate algorithms apart from each other in terms of optimization performance~\cite{benchmark-limitation}. Recent studies~\cite{ma2025toward,metabox-v2} reveal that such design bias may not keeps pace with the training\&evaluation objectives of MetaBBO, i.e., generalization toward unseen problems. One possible solution is deriving realistic instances directly from real-world application~\cite{realistic-bench}. However, collecting and formulating realistic problems require deep expertise. More importantly, the corresponding expensive simulation costs makes this solution impractical~\cite{GP-BBOB-1}. A more promising way is generating synthetic problem under proper diversity measurement~\cite{GP-BBOB-3,ma-bbob}, however, to the best of our knowledge, related researches are still very limited. 


To address the aforementioned limitations, we propose a novel generation framework that discovers diverse synthetic optimization problems through \textbf{\underline{l}}atent \textbf{\underline{s}}pace \textbf{\underline{r}}everse \textbf{\underline{e}}ngineering, termed as \textbf{LSRE}, to advance the benchmark development in BBO and MetaBBO fields. Specifically, LSRE comprises following key designs: 1) an instance space analysis~\cite{isa-smith2023instance} method that trains an autoencoder to map the Exploratory Landscape Analysis~(ELA)~\cite{ela} features of problem instances into 2-dimensional latent space; 2) an expressive symbol set including operands and operators that spans a comprehensive functional instance space; 3) a genetic programming method that effectively searches for function instance within the latent space, with an enhanced local search strategy and parallel acceleration. By grid-sampling 256 instance points from the latent space, we repeatedly call LSRE to search optimal function formula for each of them, resulting in a collection of 256 synthetic problem instances \textbf{Diverse-BBO} with adequate diversity. We validate the diversity advantage of our method against existing baseline benchmarks through large-scale MetaBBO benchmarking. The comprehensive experimental results demonstrate the effectiveness of our LSRE framework and the superiority of the generated Diverse-BBO to baseline benchmarks. We now summarize the core contributions of this paper: 
\begin{itemize}
    \item \emph{Novelty}: our paper is the first study on how to prepare high generalization potential training data for MetaBBO.
    \item \emph{Methodology}: a novel GP framework \textbf{LSRE} for generating diversified synthetic optimization problems.
    \item \emph{Benchmark}: instances generate by LSRE constitute a novel benchmark \textbf{Diverse-BBO} which could be used for development of either BBO or MetaBBO approaches.
\end{itemize}

\section{Related Works}
\subsection{Meta-Black-Box Optimization.} 
To automate the manual design and configuration of BBO algorithms, MetaBBO researches propose meta-learning algorithm design policy typically by a bi-level optimization framework (as illustrated in Fig.~\ref{fig:intro}): 1) At the meta-level, an algorithm design policy $\pi_\theta$ outputs a desired design $\omega$ for low-level optimizer $\mathcal{A}$ base on low-level optimization state $s$; 2) At the lower level, the BBO optimizer $\mathcal{A}$ is employed to optimize problem $p$ sampled from distribution $\mathcal{P}$ with $\omega$. The corresponding performance gains $r$ in low-level optimization are collected as meta-objective $ J(\theta) = \mathbb{E}_{p \in \mathcal{P}} \left[ \sum_{t=1}^{T} r_{t} \right]$, where $T$ is optimization horizon. $\pi_\theta$ is then meta-trained to maximize the meta-objective.   




Existing MetaBBO approaches can be divided into several lines according to their specific algorithm design tasks: 1) Algorithm Selection~(AS), where predefined algorithms/operators are selected by the meta-level policy to solve given problems. Early researches explored selecting operators within DE frameworks~\cite{DEDDQN,DEDQN}, and more recent approaches have investigated dynamic switching between multiple advanced optimizers along the optimization process~\cite{RLDAS}; 2) Algorithm Configuration~(AC), where given an existing optimizer, the meta-level policy dictates desired parameter values dynamically along the optimization process~\cite{LDE,metabbo-ne-2,GLEET, ConfigX}; 
3) Algorithm Generation~(AG), where the meta-level policy constructs both individual components and the entire workflow to synthesize novel BBO algorithms through algorithm workflow composition~\cite{ALDes}, novel update rules generation~\cite{Symbol} and language modelling~\cite{LLaMaCo,eoh,metabbo-icl-5}. In addition, sutdies on MetaBBO have also been extended to automated feature learning~\cite{neurela,RLDE-AFL}, efficient offline learning~\cite{q-mamba,Surr-RLDE} and direct solution manipulation~\cite{GLHF,B2OPT}. 

\subsection{Existing Benchmark Functions.}
Existing MetaBBO methods typically adopt synthetic problems from well-established BBO benchmarks~(eg., CoCo-BBOB~\cite{bbob2010}, IOHProfiler~\cite{iohprofiler}, IEEE CEC Competition series~\cite{cec2021}) and platforms~(eg., MetaBox-v1/v2~\cite{metabox,metabox-v2}). While these benchmark functions have been widely recognized and used for evaluating BBO methods, they are primarily hand-crafted and are collected with inherent biases~\cite{benchmark-limitation-2}, aimed at creating performance distinctions between optimization algorithms~\cite{benchmark-limitation}. The existing benchmarks' design bias and lack of diversity may constrain the MetaBBO to learn generalizable design policies that excel at not only the training problem distribution but also unseen problems with various landscape features~\cite{ma2025toward,neurela}. Constructing and collecting realistic problems~\cite{realistic-bench,proteindocking} could be a potential solution, but it requires deep expertise and is further limited by simulation costs~\cite{GP-BBOB-1}, ultimately rendering the solution impractical. 


Generating diverse synthetic problems might be more promising, yet several challenges persist: 1) The random function generator~\cite{GP-BBOB-3} lacks control over the characteristics and diversity of generated problems; 2) A recent line of works explore the possibility of augmenting new problem instances through affine combination of existing instances in BBO benchmark~\cite{ig&dim-reduction-dietrich2022increasing}. A representative method is MA-BBOB~\cite{ma-bbob}, which claimed achieved efficient and fine-grained problem generation. However, a key limitation is such approaches might restrict themselves within the selected composition functions. 3) In the light of symbolic regression~\cite{SR}, some exploratory studies leveraged symbolic discovery tool such as Genetic Programming~(GP)~\cite{GP-SR} to generate diversified instances by searching corresponding mathematical symbolic formulas~\cite{GP-BBOB-1,GP-BBOB-2}. However, such researches are still limited. More importantly, given the inherent expensive running of GP and the design bottlenecks in these methods, their actual generation effectiveness is still unsatisfactory. In this paper, we propose LSRE to address the limitations in the third line of work in both effectiveness and efficiency side. On the one hand, we incorporate LSRE with several novel designs: 1) Effective latent instance space analysis through autoencoder; 2) Expressive symbol set enabling both efficient generation and flexible operations; 3) Cross-dimensional local search strategy to further enhance the GP searching. On the other hand, we introduce high-performance distributed acceleration techniques to facilitate efficient GP running.

\section{Methodology}

\subsection{Latent Instance Space Analysis}\label{sec:lisa}
To align with practical usage of potential users, one of the most important issues in our framework is to pre-define a target problem range $\mathcal{D}$ as the basis for constructing corresponding the latent instance space. While interests of different users may vary, in this paper we primarily select CoCo-BBOB~\cite{bbob2010} to serve as $\mathcal{D}$ due to its ease of acquisition and long-standing reputation within BBO evaluation domain. We have to note that our framework can be easily extended to other user-specific problem ranges. We leave step-by-step instructions for such cases at Appendix A.1\footnote{``Appendix.pdf'' in our codebase.}. Given the 24 basic functions ($f_1 \sim f_{24}$) in CoCo-BBOB, we first augment these functions with five different problem dimensions $[2,5,10,30,50]$, resulting in 120 functions. Next, we further generate a sufficient number of instances by applying random rotation and shift transformations to these 120 functions:
\begin{equation}
    f^{\prime}(x) = f(\mathbf{R}^T(x-s))
\end{equation}

where $f$ denotes the original function and $f^{\prime}$ the transformed one, $x$ is the decision variables, $\mathbf{R}$ and $s$ are random rotation matrix and shift vector, respectively. We randomly generate 270 instances for each of the 120 functions following the equation above, resulting in 32,400 instances as selected $\mathcal{D}$, which is used in following sections. 
\begin{figure}[t]
    \centering
    \includegraphics[width=0.85\linewidth]{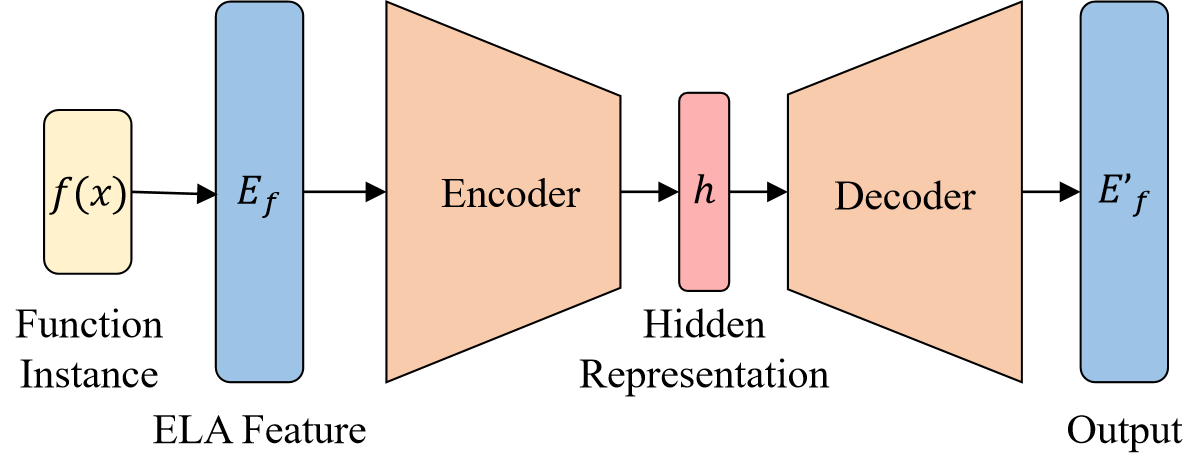}
    \caption{Workflow of Latent Instance Space Analysis.}
    \label{fig:AE}
\end{figure}
We construct latent instance space of $\mathcal{D}$ by combining the Exploratory Landscape Analysis~(ELA)~\cite{ela} and popular dimension reduction strategy autoencoder~\cite{AE-0}. We illustrate the workflow of such latent instance space analysis in Fig.~\ref{fig:AE}, and briefly introduce how these two techniques cooperate with each other:

\textbf{ELA.} It is a widely used tool for instance space analysis~\cite{bbob-ela-analysis&dim-reduction-vskvorc2020understanding} of BBO problems. Groups of analytic or statistical features, such as convexity, global structure and meta-model fitting are computed in ELA, which has been proven informative in algorithm selection~\cite{ela4as}. Given a BBO problem instance $f \in \mathcal{D}$, we obtain its ELA features vector $E_f$ using open-source ELA package: pflacco\footnote{https://github.com/Reiyan/pflacco}. Specifically, we select a subset of representative and computationally efficient features as $E_f$. The selection standard and concrete feature list are provided in Appendix A.2 . 

\textbf{Autoencoder.} An autoencoder comprises two network components: the encoder $W_\theta$ and the decoder $W_\phi$. The input to $W_\theta$ is a high-dimensional feature vector, and it then maps this feature vector into a low-dimensional hidden space. Given the hidden feature output by $W_\theta$, the decoder $W_\phi$ maps it back to the original high-dimensional feature space. In our case, the high-dimensional space spans the ELA features space of the selected problem range $\mathcal{D}$, and the hidden space is a 2-dimensional space denoted as $\mathcal{H}$. The training of our autoencoder follows the normal reconstruction loss function used in existing literature. Given a BBO problem instance $f \in \mathcal{D}$, we first compute its ELA features $E_f$, the reconstructed feature $E^\prime_f$ is obtained by feeding $E_f$ into $W_\theta$ and $W_\phi$. Then the reconstruction loss is formulated as:
\begin{equation}
    \mathcal{L}(\theta,\phi) = \frac{1}{2}||E_f - E^\prime_f||_2^2, \quad f\sim \mathcal{D}
\end{equation}
We leave the settings of the autoencoder such as its architecture and corresponding training protocol in Appendix A.3. Once the training of the autoencoder finishes, we could obtain latent representation $h \in \mathbb{R}^2$ of a BBO problem instance through the encoder. More importantly, the well-trained model allows us to measure the diversity of problem instances in a 2-dimensional space, which further facilitates uniform instance sampling from this hidden space, since directly sampling from high-dimensional ELA space is neither efficient nor practical. It is worth mentioning that principal component analysis~(PCA)~\cite{pca} is also a popular dimension reduction approach widely adopted in machine learning community. The reason behind our autoencoder choice is that: 1) PCA shows effectiveness in cases where features hold linear correlation assumption~\cite{AE-1,AE}, however, different feature groups in ELA are independently computed; 2) We have conducted a preliminary study on using PCA for ELA feature reduction, the results show that the first two principle components of certain portion of instances in $\mathcal{D}$ can not surpass 51\% importance threshold. We also provide an ablation study in Sec.~\ref{sec:ablation}.     


\subsection{Genetic Programming Search}\label{sec:gp}
Through the latent instance space analysis, we now obtain a 2-dimensional latent instance space $\mathcal{H}$. To generate synthetic BBO problem instance from $\mathcal{H}$, several challenges still remain. One of them is how we model a proximal function space to represent or formulate mathematical functions of optimization problems as computer-understandable language. To address this, we adopt a symbolic tree representation for mathematical function, which has proven to be robust and effective in symbolic regression domain~\cite{SR}. To make it a comprehensive proximal space, we design a flexible symbol set containing diverse operators and operands, which facilitates accurate and efficient proximal representation. Another key challenge is how we reverse engineer a proximal representation of a BBO problem from the latent space $\mathcal{H}$. Genetic Programming~(GP)~\cite{GP-SR} is regarded as an effective way to discover functions from symbolic function spaces~\cite{GP-SR-1}. In this paper, we propose a novel GP framework to search for the fittest function formula given a latent instance point in $\mathcal{H}$. This includes how we define the searching objective and how we enhance naive GP in both effectiveness and efficiency sides. We present the pseudocode of the GP search in Alg.~\ref{alg:algorithm} and elaborate our designs in the following sections.

\subsubsection{Proximal Symbolic Space}
\begin{figure}[t]
    \centering
    \includegraphics[width=0.7\linewidth]{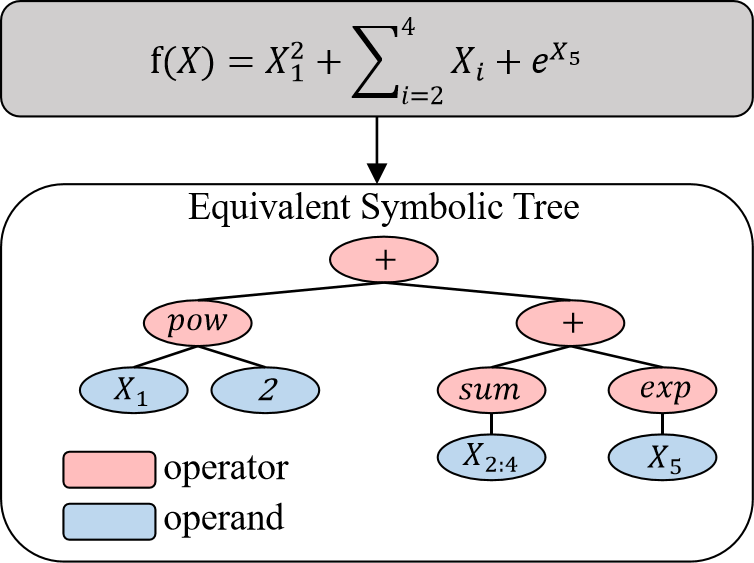}
    \caption{A proximal symbolic tree example.}
    \label{fig:SR}
\end{figure}
We propose a comprehensive symbol set comprising flexible operators and operands that spans a proximal symbolic space in regard of single-objective continuous optimization problems. Specifically, the symbolic set comprises following symbols:
\begin{itemize}
    \item \emph{Operator}: 15 operators including \verb|add|, \verb|sub|, \verb|mul|, \verb|div|, \verb|neg|, \verb|pow|, \verb|sin|, \verb|cos|, \verb|tanh|, \verb|exp|, \verb|log|, \verb|sqrt|, \verb|abs|, \verb|sum|, and \verb|mean|. Except \verb|sum| and \verb|mean|, the listed operators are all unary/binary operators commonly used in symbolic regression research. We replenish them with \verb|sum| and \verb|mean| as additional aggregation operation, which could simplify multi-addend structure in symbolic tree, so as to enhance the searching efficiency of GP.
    \item \emph{Operand}: three operands including \verb|X|, \verb|X[i:j]| and \verb|C|. \verb|X| denotes the vector of decision variables. \verb|C| denotes constant values or vectors. \verb|X[i:j]| denotes indexed decision variables, e.g., \verb|X[2:4]| indexes the second to fourth decision variables while \verb|X[2:2]| denotes the second one. It is worthy to mention that compared with symbolic set design in existing literature which only involve \verb|X| and \verb|C|, we add \verb|X[i:j]| to achieve fine-grained generation.
\end{itemize}
With the comprehensive symbol set, we can sample diverse optimization problems by constructing their equivalent symbolic trees from the corresponding symbolic space~(see Fig.~\ref{fig:SR}). In the proposed GP search, we represent each candidate function instance by its symbolic tree and use tree-based evolutionary operators to search for target function instance. We leave a more detailed discussion of the symbol set at Appendix A.4. 

\subsubsection{Searching Objective} The objective of our proposed GP search is quite straightforward. Considering we want to reverse engineer~(i.e.,search for) the mathematical formula from a given instance point $h\in \mathcal{H}$, where $\mathcal{H}$ is the 2-dimensional latent instance space we constructed. We use $\tau$ to denote any legal symbolic tree constructed from our symbol set, and use $E_\tau$ to denote the landscape features of the corresponding problem instance. Then the searching objective is to minimize a score function $obj(\tau) = \frac{1}{2}||W_\theta(E_\tau) - h||_2$, where $W_\theta$ is the encoder of the autoencoder in latent instance space analysis, and to find an optimal $\tau^*$ such that:
\begin{equation}\label{eq:obj}
    \tau^* = \arg\min\limits_{\tau} obj(\tau)
\end{equation}

\subsubsection{Searching Strategy} We aim to reverse engineer a synthetic problem instance by finding its optimal proximal representation $\tau^*$ in Eq.~\ref{eq:obj}. In LSRE, we adopt \emph{gplearn}\footnote{https://github.com/trevorstephens/gplearn} to achieve this goal, which is an effective and robust GP library. We initialize a \emph{gplearn} process with our proposed symbol set and searching objective. 

The pseudocode of the GP search in LSRE is provided in Alg.~\ref{alg:algorithm}. In lines 2-3, we first randomly initialize a population of proximal tree representations $\{\tau_i\}_{i=1}^N$ by sampling from the symbolic space spanned by our proposed symbol set, and then evaluate them to attain their objective values $obj(\cdot)$. Then we begin an iterative searching process in lines 4-13. For each generation $g$, the solution population first undergoes a roulette\_reproduction operation (line 6). It is a built-in reproduction strategy of \emph{gplearn}, which randomly assign one of five default mutation/crossover operators for each tree individual, by a predetermined operator selection preference. After roulette\_reproduction operation, an offspring population $\{\tau_i^\prime\}_{i=1}^N$ is obtained. In line 7, we propose a novel plug-and-play cross-dimensional local search strategy which can be easily integrated into the GP process to improve the searching accuracy. The motivation behind this local search strategy is that: although we could search for optimal function formulation, the dimension of the decision variables is not addressed. Several recent studies~\cite{dimension-ela-1, dimension-ela-2} reveal that even two function instances share mathematical formulation, their ELA features may vary due to different problem dimensions. To address this, in our proposed local search strategy, for each offspring $\tau_i\prime$, we instantiate it to 2D, 5D and 10D function instances and compare their objective values, then the best objective value is regarded as the objective value of this offspring. We provide effectiveness validation in following ablation studies. After a total of $G$ generations GP searching, we obtain the best solution $\tau^*$ found so far (line 14) as well as its best-performing problem dimension.    

\begin{algorithm}[t]
\caption{Genetic Programming Search in LSRE}
\label{alg:algorithm}
\textbf{Input}: population size \emph{N}; budget \emph{G}; searching target \emph{h}.\\
\textbf{Output}: best instance $\tau^*$ found ever.
\begin{algorithmic}[1] 
\State Let $g=1$
\State Initialize a tree population $\{\tau_1,...,\tau_N\}$
\State Evaluate the tree population $\{obj(\tau_1),...,obj(\tau_N)\}$
\While{$g \leq G$}
\For{each tree individual $\tau_i$}
\State $\tau_i^\prime \leftarrow \text{roulette\_reproduction}(\tau_i)$
\State $\tau_i^{\prime\prime} \leftarrow \text{cross\_dimension\_local\_search}(\tau_i^\prime)$
\If {$obj(\tau_i^{\prime\prime}) \le obj(\tau)$}
\State $\tau_i \leftarrow \tau_i^{\prime\prime}$.
\EndIf
\EndFor
\State $g = g+1$
\EndWhile
\State $\tau^* \leftarrow \arg\min\limits_{\tau_i} obj(\tau_i)$
\State \textbf{return} $\tau^*$
\end{algorithmic}
\end{algorithm}

\subsection{Diverse-BBO}\label{sec:diversebbo}
\subsubsection{Overall Workflow} With the latent instance space and the GP searching strategy at hand, we now elaborate on how LSRE generates uniform-diversity function instances for Diverse-BBO. We first demarcate a grid area in the 2-dimensional latent space, i.e., $x \in [-B,B]$ and $y \in [-B,B]$, which could cover the target problem range $\mathcal{D}$ we constructed in Sec.~\ref{sec:lisa}. Then we uniformly sample $K \times K$ latent instance points (point matrix) from this grid area, where $K$ controls the sampling granularity. At last, for each sampled instance point, we call our GP search to reverse engineer -corresponding function formulation. In this paper, we set $B=1$ and $K=16$ to obtain a problem set of 256 diversified instances, which we denote as Diverse-BBO. Refer to ```DiverseBBO.pdf'' in our codebase for their details.  

\subsubsection{Distributed Acceleration}
To obtain Diverse-BBO is challenging considering that: 1) we have to repeatedly call the GP process for 256 times; 2) Within each GP process, evaluating the tree population is also a tedious and time-consuming task due to repeated objective computation. The overall complexity is hence $\mathcal{O}[M*G*(N^2+N*L^2)]$, where M is the number of functions we need to generate (256), G is the GP generations (50), N and L are population size and tree node size respectively. To accelerate LSRE, we further introduce Ray\footnote{https://github.com/ray-project/ray} to first distribute 256 GP processes on independent CPU cores and then trigger second-level parallel evaluation within each GP process. By doing this, we significantly reduce the overall complexity to $\mathcal{O}[G*L^2)]$. We leave the hyper-parameters including their selection standards and the detailed design philosophy behind LSRE's components at Appendix A.5 and Appendix A.6. 

\begin{figure}[t]
    \centering
    \includegraphics[width=1\linewidth]{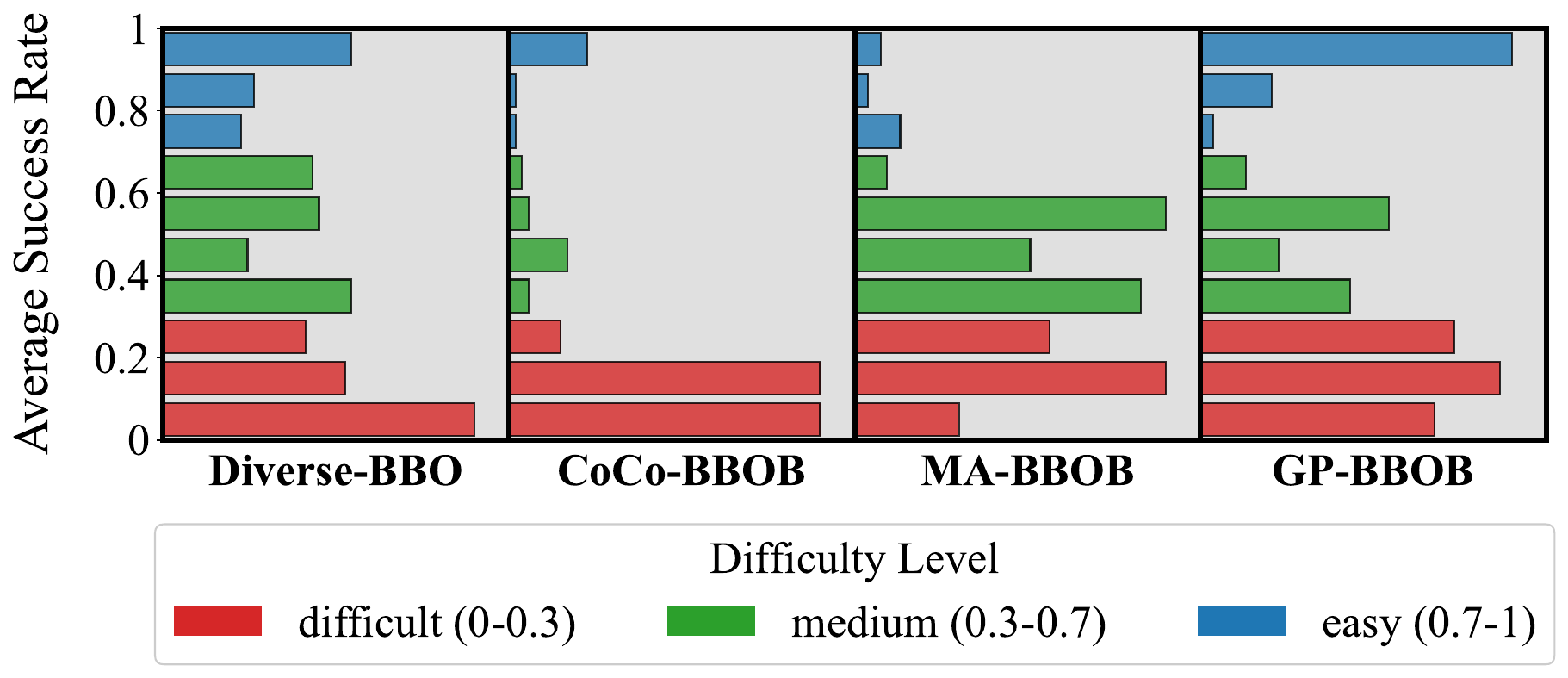}
    \caption{Average Success Rate Distribution}
    \label{fig:avg-sr-pic}
\end{figure}

\begin{table*}[ht!]
    \centering
\resizebox{0.8\linewidth}{!}{
\begin{tabular}{c|cc|cc|cc|cc|cc}
        \hline
        \multicolumn{11}{c}{Test Set: $\mathbb{D}_{synthetic}$} \\
        \hline 
         & \multicolumn{2}{c|}{DEDDQN} & \multicolumn{2}{c|}{LDE} & \multicolumn{2}{c|}{SYMBOL} & \multicolumn{2}{c|}{GLEET}& \multicolumn{2}{c}{Average} \\
        & Mean$\uparrow$($\pm$Std) & Rank &Mean$\uparrow$($\pm$Std) & Rank &Mean$\uparrow$($\pm$Std) & Rank &Mean$\uparrow$($\pm$Std) & Rank &Mean$\uparrow$($\pm$Std) & Rank \\
    \hline
    \hline
    Diverse-BBO & 
    \begin{tabular}[c]{@{}c@{}}\textbf{8.154e-01} \\ \textbf{ ($\pm$5.219e-02)} \end{tabular} & 
    \begin{tabular}[c]{@{}c@{}} \textbf{1}\end{tabular} & 
    \begin{tabular}[c]{@{}c@{}} \textbf{8.315e-01} \\  \textbf{($\pm$3.136e-02)}\end{tabular} & 
    \begin{tabular}[c]{@{}c@{}}\textbf{1}\end{tabular} & 
    \begin{tabular}[c]{@{}c@{}} \textbf{7.346e-01 }\\ \textbf{($\pm$9.217e-02)}\end{tabular} & 
    \begin{tabular}[c]{@{}c@{}} \textbf{1}\end{tabular} & 
    \begin{tabular}[c]{@{}c@{}} 8.106e-01\\  ($\pm$5.983e-02)\end{tabular} & 
    \begin{tabular}[c]{@{}c@{}} 2\end{tabular} & 
    \begin{tabular}[c]{@{}c@{}}  \textbf{7.980e-01}\\ \textbf{ ($\pm$5.889e-02)}\end{tabular} & 
    \begin{tabular}[c]{@{}c@{}}\textbf{1.25} \end{tabular} \\ 
    \hline
    CoCo-BBOB & 
    \begin{tabular}[c]{@{}c@{}} 8.106e-01\\  ($\pm$5.217e-02)\end{tabular} & 
    \begin{tabular}[c]{@{}c@{}} 2\end{tabular} & 
    \begin{tabular}[c]{@{}c@{}} 8.271e-01 \\ ($\pm$3.128e-02)\end{tabular} & 
    \begin{tabular}[c]{@{}c@{}} 2\end{tabular} & 
    \begin{tabular}[c]{@{}c@{}} 7.319e-01\\  ($\pm$7.363e-02)\end{tabular} & 
    \begin{tabular}[c]{@{}c@{}} 2\end{tabular} & 
    \begin{tabular}[c]{@{}c@{}} 8.058e-01 \\($\pm$4.146e-02)\end{tabular} & 
    \begin{tabular}[c]{@{}c@{}} 4\end{tabular} & 
    \begin{tabular}[c]{@{}c@{}}  7.939e-01\\($\pm$4.963e-02)\end{tabular} & 
    \begin{tabular}[c]{@{}c@{}} 2.5\end{tabular} \\ 
    \hline
    MA-BBOB & 
    \begin{tabular}[c]{@{}c@{}} 8.006e-01 \\ ($\pm$5.133e-02)\end{tabular} & 
    \begin{tabular}[c]{@{}c@{}}4 \end{tabular} & 
    \begin{tabular}[c]{@{}c@{}} 8.232e-01 \\ ($\pm$3.741e-02)\end{tabular} & 
    \begin{tabular}[c]{@{}c@{}}3 \end{tabular} & 
    \begin{tabular}[c]{@{}c@{}} 7.021e-01 \\ ($\pm$9.149e-02)\end{tabular} & 
    \begin{tabular}[c]{@{}c@{}} 3\end{tabular} & 
    \begin{tabular}[c]{@{}c@{}} \textbf{8.112e-01} \\  \textbf{($\pm$4.664e-02)}\end{tabular} & 
    \begin{tabular}[c]{@{}c@{}} \textbf{1}\end{tabular} & 
    \begin{tabular}[c]{@{}c@{}}  7.843e-01\\($\pm$5.672e-02)\end{tabular} & 
    \begin{tabular}[c]{@{}c@{}} 2.75\end{tabular} \\
    \hline
    GP-BBOB & 
    \begin{tabular}[c]{@{}c@{}}8.045e-01\\($\pm$5.378e-02)\end{tabular} & 
    \begin{tabular}[c]{@{}c@{}} 3\end{tabular} & 
    \begin{tabular}[c]{@{}c@{}} 8.223e-01 \\($\pm$3.938e-02)\end{tabular} & 
    \begin{tabular}[c]{@{}c@{}} 4\end{tabular} & 
    \begin{tabular}[c]{@{}c@{}} 6.907e-01\\  ($\pm$7.625e-02)\end{tabular} & 
    \begin{tabular}[c]{@{}c@{}} 4\end{tabular} & 
    \begin{tabular}[c]{@{}c@{}} 8.088e-01 \\($\pm$4.827e-02)\end{tabular} & 
    \begin{tabular}[c]{@{}c@{}} 3\end{tabular} & 
    \begin{tabular}[c]{@{}c@{}}  7.816e-01\\($\pm$5.442e-02)\end{tabular} & 
    \begin{tabular}[c]{@{}c@{}} 3.5\end{tabular} \\
    \hline
    \end{tabular}
    }
    \caption{Generalization performance of MetaBBO when being trained on different baseline benchamrks.}
    \label{tab:generalization-result}
\end{table*}

\section{Experimental Analysis}
We validate the effectiveness of our proposed LSRE through comparing Diverse-BBO generated by LSRE with representative baseline benchmarks. In specific, we consider following research questions. \textbf{RQ1: }Can Diverse-BBO show more diversity compared to other synthetic BBO benchmarks? \textbf{RQ2: }Can MetaBBO approaches benefit from such diversity in terms of meta-learned generalization? \textbf{RQ3: }How does our proposed novel designs contribute to LSRE? Below we first introduce all baselines and then discuss RQ1$\sim$RQ3 in the following sections. The baselines are: 1) Diverse-BBO: it includes 256 diversified synthetic problems generated by our LSRE, the problem dimensions for these instances are 2, 5, and 10; 2) CoCo-BBOB~\cite{bbob2010}: it comprises 24 basis synthetic functions, which we used to generate 256 10-dimensional instances by transforming them via random rotation and shift; 3) MA-BBOB~\cite{ma-bbob}: it uses 24 basis fucntions in CoCo-BBOB as component functions and generates new instance by affine combination of them. We use MA-BBOB to generate 256 10-dimensional instances as a baseline; 4) GP-BBOB~\cite{GP-BBOB-1}: the first framework that uses GP for BBO instance generation. Our LSRE differs with this method in instance space analysis, symbol set construction and searching strategy. We use it to generate 256 instances as a baseline.

\subsection{Diversity Comparison (RQ1)}

We first show how we measure the diversity of a benchmark. Existing literature of BBO benchmarks barely provide explicit definition of diversity for the testing problem instances included. However, some existed papers indeed provide indirect measurements that reflect benchmark diversity through algorithm performance distribution~\cite{diversity-1}. Following this idea, we propose a simple and intuitive average success rate distribution~(ASRD) test to reflect the diversity of a benchmark through the problem difficulty distribution. In ASRD test, for a given benchmark that includes $N$ problem instances $f_1 \sim f_N$, we first construct a BBO optimizer pool that comprises $M$ representative optimizers $\mathcal{A}_1 \sim \mathcal{A}_M$. For any pair of problem instance $f_i$ and optimizer $\mathcal{A}_j$, we use $\mathcal{A}_j$ to optimize $f_i$ for 51 independent runs and record the portion of successful runs~(runs that achieve optimal on $f_i$) as success rate $SR_{i,j}$. Then the ASRD for the given benchmark is summarized as the statistical histogram of all success rate results $\{\{SR_{i,j}\}_{i=1}^N\}_{j=1}^M$.  

In this experiment, we conduct ASRD test for all four baselines, the optimizer pool we construct include several representative optimizers such as DE, PSO, CMA-ES etc.. For more detailed experiment protocol, we kindly request readers refer to Appendix B.1. We visualize the test results in Fig.~\ref{fig:avg-sr-pic}, which show that: 1) \textbf{Diverse-BBO v.s. CoCo-BBOB}: derived from the latent space of CoCo-BBOB, our Diverse-BBO achieves preferred problem diversity through fine-grained generation of LSRE; 2) \textbf{Diverse-BBO v.s. MA-BBOB}: certain diversity improvement can be observed in MA-BBOB compared to CoCo-BBOB, however, the affine combination in MA-BBOB still fails to achieve controllable uniform diversity distribution; 3) \textbf{Diverse-BBO v.s. GP-BBOB}: sharing the same GP-based generation backbone, the novel designs we proposed in LSRE result in better generation quality, which we will discuss more in the subsequent sections.

\subsection{Generalization Test (RQ2)}
Another key research question is whether the diversity of a BBO benchmark truly provides convenience for the generalization of MetaBBO. To this end, we conduct a large-scale validation experiment in this section. 
\subsubsection{Settings}
For the four baseline benchmarks, we use them for training MetaBBO approaches and then observe their generalization performance. This results in three experiment protocols we have to clarify here. First, for each baseline, we randomly split it into 75\% training instances and 25\% testing instances. We then collect all testing instances of the four benchmarks into one test set, termed as $\mathbb{D}_{synthetic}$. Second, we further prepare another three test sets $\mathbb{D}_{hpo}$, $\mathbb{D}_{uav}$ and $\mathbb{D}_{protein}$, which denote complex realistic BBO problems: hyper-parameter optimization~\cite{hpob}, uav planning~\cite{uav} and protein docking~\cite{proteindocking}. Third, we select four representative MetaBBO approaches from existing literature: DEDDQN~\cite{DEDDQN}, LDE~\cite{LDE}, SYMBOL~\cite{Symbol} and GLEET~\cite{GLEET}. These baselines cover different methodology categories such as those for operator selection~(DEDDQN), for algorithm configuration~(LDE \& GLEET) and for algorithm generation~(SYMBOL). We use their default settings for training and testing according to the original papers. More details about these protocols are provided at Appendix B.2. 

\begin{figure}[t]
    \centering
    \includegraphics[width=0.95\linewidth]{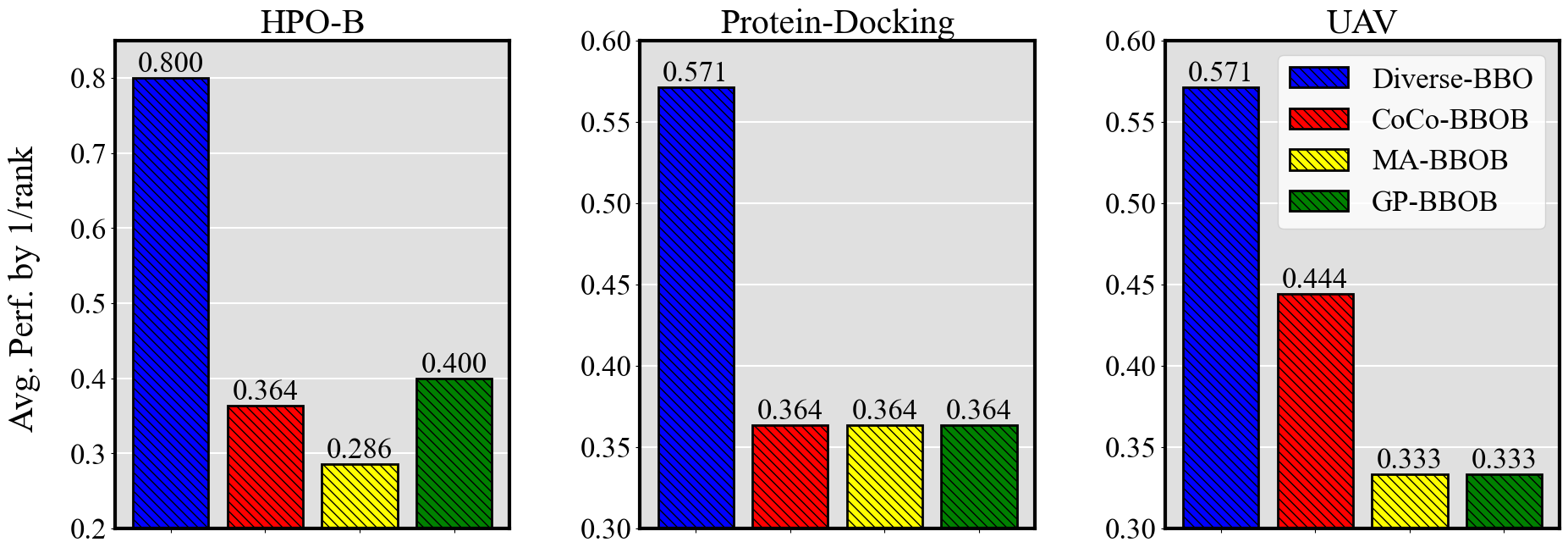}
    \caption{Generalization performance on realistic problems.}
    \label{fig:realistic}
\end{figure}

\subsubsection{Synthetic Test} 
We train the four MetaBBO approaches on the four baseline benchmarks' train sets respectively, with the same random seed to ensure the comparison fairness. We then test these 16 (4 MetaBBOs, 4 train sets) trained models on $\mathbb{D}_{synthetic}$, with 51 independent runs on each problem instance. We adopt the average normalized optimization performance metrics (larger is better) widely used in MetaBBO approaches~\cite{ConfigX,neurela,GLHF} in the test. The testing results are presented in Table~\ref{tab:generalization-result}, where the ranks of baseline benchmarks are annotated alongside the performance metric, and the average ranks and performance across diverse MetaBBOs are also presented in the rightmost column. The results show that generally the diversity advantage of our Diverse-BBO against the other baselines facilitates the generalization performance of various MetaBBO approaches. It only slightly underperforms MA-BBOB with large overlapping. An interesting observation is that though MA-BBOB and GP-BBOB have been claimed useful for evaluating traditional BBO optimizer, they show an opposite effect in MetaBBO domain, which anticipates further investigation in the future.

\subsubsection{Realistic Test} 
\begin{figure}[t]
    \centering
    \includegraphics[width=0.75\linewidth]{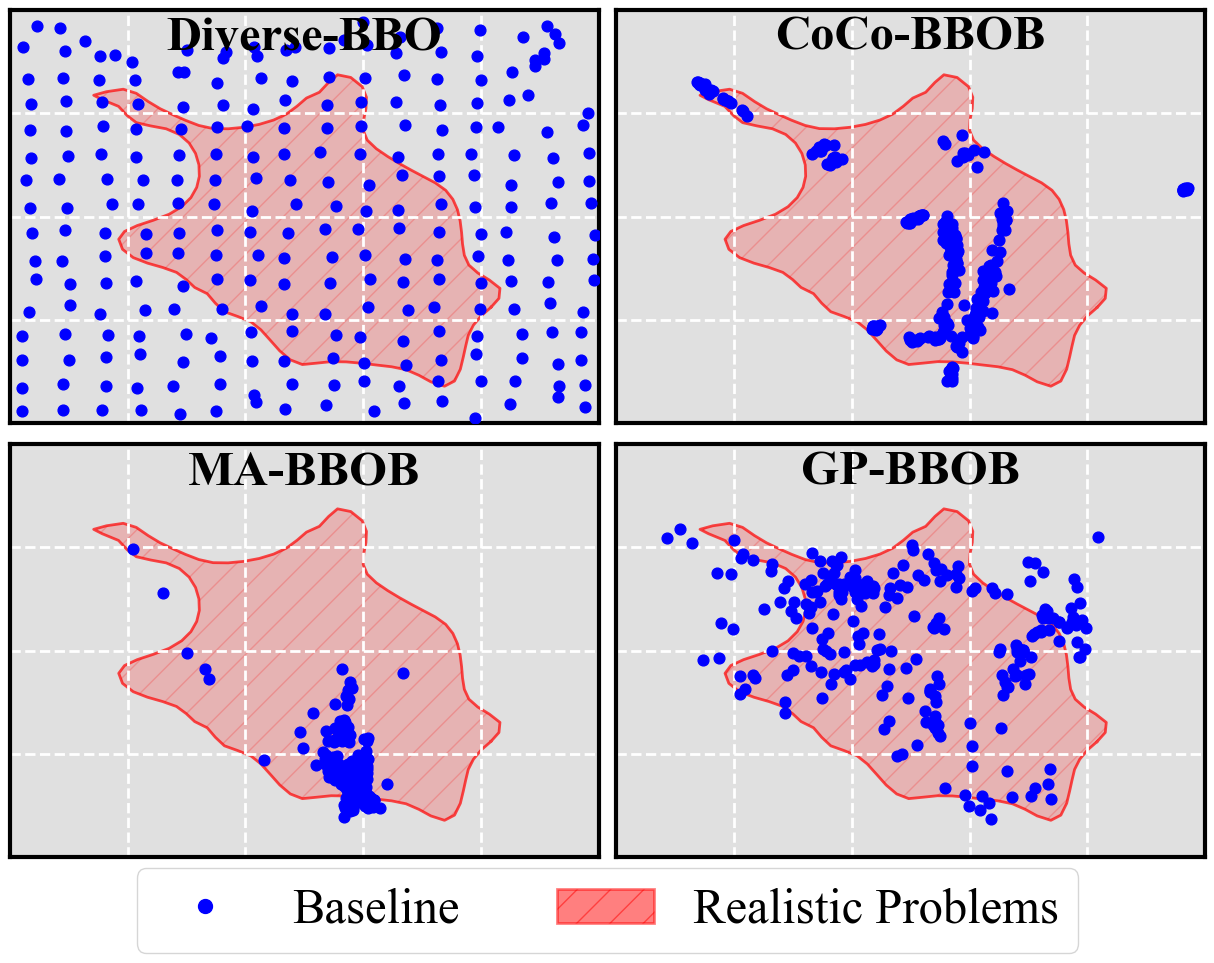}
    \caption{Generalization interpretation of baselines.}
    \label{fig:four-bench-in-latent-space}
\end{figure}

According to the insights obtained from recent MetaBBO-related  researches, out-of-distribution generalization, especially MetaBBO's zero-shot performance on unseen realistic problems should be thoroughly evaluated~\cite{metabox-v2}. To this end, in this section we set a very high bar for the four baseline benchmarks. We test the four MetaBBO approaches trained previously on three realistic problem sets $\mathbb{D}_{hpo}$, $\mathbb{D}_{uav}$ and $\mathbb{D}_{protein}$. We illustrate in Fig.~\ref{fig:realistic} the average performance gains of baseline benchmarks across the four MetaBBO approaches, which is computed as $\frac{1}{rank}$. The results further demonstrate the diversity advantage of our Diverse-BBO, indicating the effectiveness of LSRE in generating high-quality training instances for MetaBBO. To further interpret the observed superiority, we provide an intuitive visualization in Fig.~\ref{fig:four-bench-in-latent-space} where the shaded red area denotes the coverage of all instances of the three realistic sets in the latent space $\mathcal{H}$ we constructed in Sec.~\ref{sec:lisa}. The blue points denotes the positions of the 256 instances of each baseline benchmark in $\mathcal{H}$. We believe results there could clearly explain the relationship between benchmark diversity and corresponding generalization potential. Through our LSRE, we achieve controllable, uniform and fine-grained instance generation within the instance space, which hence benefits training of MetaBBO.        

\begin{figure}[t]
    \centering
    \includegraphics[width=0.8\linewidth]{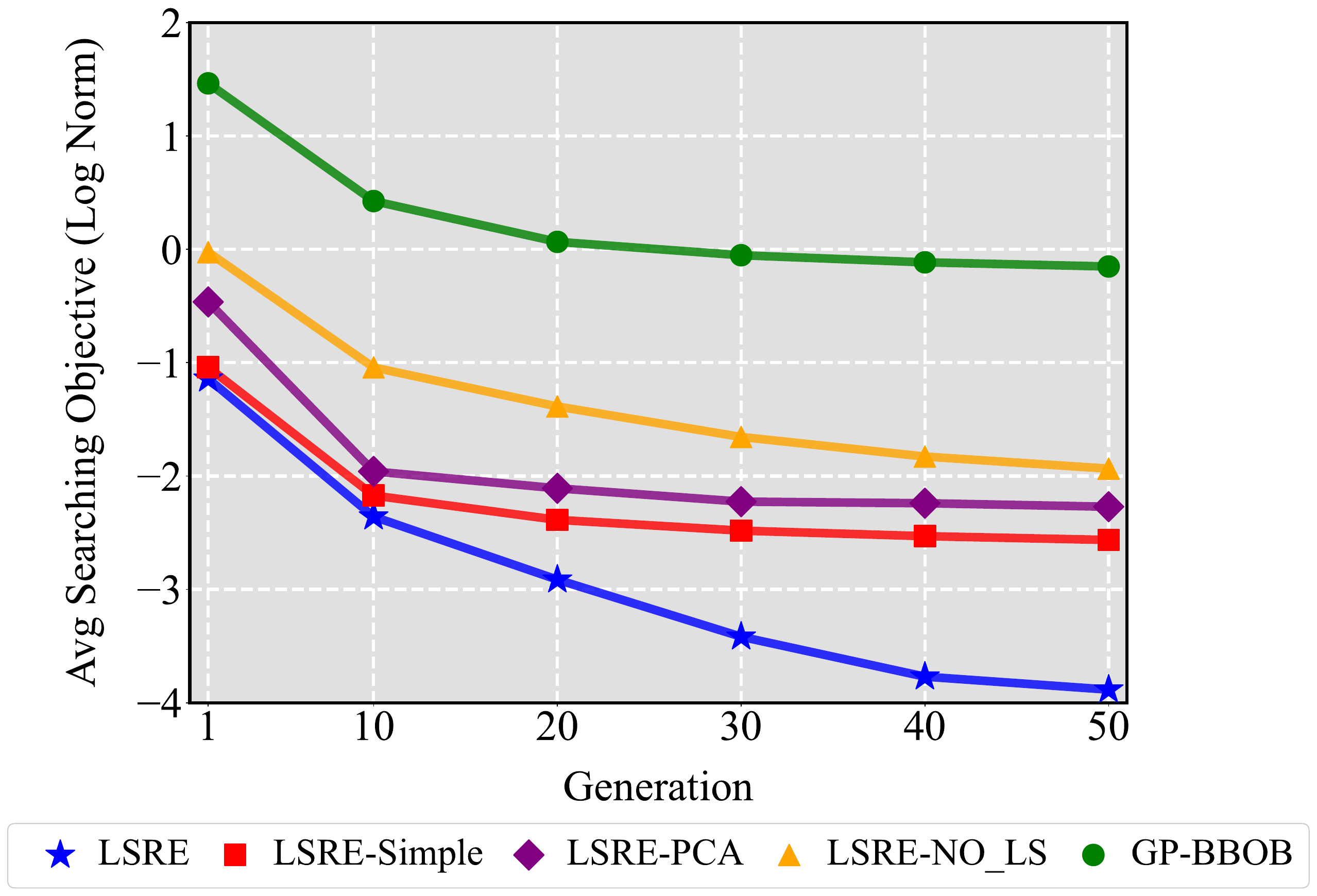}
    \caption{LSRE optimization curves when setting vary.}
    \label{fig:abalation}
\end{figure}

\subsection{Ablation Study (RQ3)}\label{sec:ablation}
We provide following ablation baselines to validate our proposed novel designs in LSRE: 1) LSRE-PCA: where we use PCA for the latent instance space analysis, selecting the most important two features for dimension reduction; 2) LSRE-Simple: where we use simpler symbol set design of existing symbolic regression research~\cite{GP-BBOB-3}, without vectorized operations; 3) LSRE-NO\_LS: where the cross-dimensional local search strategy is ablated from the GP search process. Besides, we additionally include a background baseline GP-BBOB~\cite{GP-BBOB-1}, which can be regarded as a baseline that ablates all novel designs in LSRE. Specifically, GP-BBOB share the same GP backbone as our LSRE, while it uses the simpler symbol set, PCA-based instance space analysis and naive GP search.

 We present in Fig.~\ref{fig:abalation} the optimization curves of these ablation baselines and our LSRE when searching for $\tau^*$, which are averaged across all 256 target instance points. The results not only demonstrate significant improvement LSRE achieved against GP-BBOB, but also validate that each novel design in LSRE contributes to the final performance. Specifically, while the augmented symbol set and refined autoencoder-based instance analysis provide comparably equal contribution, we observe that the cross-dimensional local search plays a more important role in LSRE. Compared to existing works that fix the problem dimension during their searching process~\cite{GP-BBOB-1,GP-BBOB-2,ma-bbob}, our finding reveals that the ELA feature space might not be smooth enough if we do not co-optimize the problem dimension. There are also several insightful and interesting findings such as the visualized comparison between PCA and autoencoder in instance space analysis, the complexity of the function instance generated by LSRE etc.. We leave them at Appendix B.3 for curious readers.

\section{Conclusions}
This paper serves as the first study to emphasize the importance of training data for Meta-Black-Box Optimization~(MetaBBO). We have identified design bias and limitations of existing BBO benchmark and related instance generation approach through empirical evaluation. To address such design bias, we propose LSRE framework that first leverages an effective autoencoder-based latent instance space analysis to establish a comprehensive searching space. LSRE then deploys GP-based instance generation~(searching) method to discover diversified synthetic problem instances. Several novel designs facilitate this process such as a replenished symbol set to enhance expressiveness for mathematical formulas and a cross-dimensional local search strategy to significantly improve the generation accuracy. We obtain several valuable conclusions from the comprehensive experiment results: 1) While existing BBO benchmarks hold long-stading reputation on evaluating traditional BBO optimizers, they may not be a proper choice for training MetaBBO; 2) The diversity of training problem set significantly impacts the generalization potential of MetaBBO; 3) The Diverse-BBO problem set generated by our LSRE framework, is at least so far, a better choice for future MetaBBO researches. We appeal more and more researchers for related studies on algorithm performance evaluation, benchmark problem diversity and generalization potential within both BBO and MetaBBO fields.        

\section{Acknowledgments}
This work was supported in part by National Natural Science Foundation of China (Grant No. 62276100), in part by the Guangdong Provincial Natural Science Foundation for Outstanding Youth Team Project (Grant No. 2024B1515040010), in part by Guangzhou Science and Technology Elite Talent Leading Program for Basic and Applied Basic Research (Grant No. SL2024A04J01361), and in part by the Fundamental Research Funds for the Central Universities (Grant No. 2025ZYGXZR027).



\bibliography{aaai2026}

\end{document}